\documentclass{bmvc2k}

\usepackage{soul}
\usepackage{url}
\usepackage[small]{caption}
\usepackage{subfigure}
\usepackage{amsmath}
\usepackage{amsthm}
\usepackage{amssymb}
\usepackage{multirow}
\usepackage{makecell}
\usepackage{booktabs}
\usepackage{algorithm}
\usepackage{algorithmic}
\usepackage{indentfirst}
\usepackage{xr-hyper}

\usepackage[capitalize,noabbrev]{cleveref}
\usepackage{optidef}
\usepackage{newfloat}
\usepackage{listings}
\usepackage{mathtools}
\usepackage{bm}
\usepackage{bbm}
\usepackage{multirow}
\usepackage{array}
\newcolumntype{H}{>{\setbox0=\hbox\bgroup}c<{\egroup}@{}}
\usepackage{tikz}
\usetikzlibrary{fit,backgrounds}

\theoremstyle{plain}

\theoremstyle{definition}

\theoremstyle{remark}

\floatstyle{ruled}
\newfloat{listing}{tb}{lst}{}
\floatname{listing}{Listing}

\newcommand{\one}[1]{\mathbbm{1}}

\newcommand{\eg}{e.g.}
\newcommand{\ie}{i.e.}
\AtBeginDocument{\hypersetup{pdfborder={0 0 0}}}

\newcommand{\inc}[1]{$_{\text{#1}\uparrow}$}

\providecommand{\eg}{\textit{e.g.}\@\xspace}
\providecommand{\ie}{\textit{i.e.}\@\xspace}

\newcommand{\Hquad}{\hspace{0.5em}} 

\definecolor{codegreen}{rgb}{0,0.6,0}
\definecolor{codegray}{rgb}{0.5,0.5,0.5}
\definecolor{codepurple}{rgb}{0.58,0,0.82}
\definecolor{backcolour}{rgb}{0.95,0.95,0.92}
\definecolor{darkgreen}{RGB}{0,152,68}

\lstdefinestyle{mystyle}{
    backgroundcolor=\color{white},   
    commentstyle=\color{codegreen},
    keywordstyle=\color{magenta},
    numberstyle=\tiny\color{codegray},
    stringstyle=\color{codepurple},
    basicstyle=\ttfamily\footnotesize,
    breakatwhitespace=false,         
    breaklines=true,                 
    captionpos=b,                    
    keepspaces=true,                 
    numbers=none,                    
    numbersep=5pt,                  
    showspaces=false,                
    showstringspaces=false,
    showtabs=false,                  
    tabsize=2
}

\newcommand{\augview}{augmented view}

\newcommand{\imageaugmentation}{image augmentation}

\newcommand{\netparams}{{\theta}}
\newcommand{\targetparams}{\xi}

\newcommand*{\eqdef}{\triangleq}
\newcommand{\shortcite}[1]{(\citeyear{#1})}

\title{A Unified Mixture-View Framework for Unsupervised Representation Learning}

\addauthor{Xiangxiang Chu}{chuxiangxiang@meituan.com}{1}
\addauthor{Xiaohang Zhan}{xiaohangzhan@outlook.com}{2}
\addauthor{Bo Zhang}{zhangbo97@meituan.com}{1}

\addinstitution{
Meituan\\
Beijing, China
}
\addinstitution{
The Chinese University of Hong Kong \\
HongKong, China
}

\runninghead{Chu et al.}{BSIM}

\def\eg{\emph{e.g}\bmvaOneDot}

\begin{document}

\maketitle

\begin{abstract}
Recent unsupervised contrastive representation learning follows a Single Instance Multi-view (SIM) paradigm where positive pairs are usually constructed with intra-image data augmentation. In this paper, we propose an effective approach called Beyond Single Instance Multi-view (BSIM). Specifically, we impose more accurate instance discrimination capability by measuring the joint similarity between two randomly sampled instances and their mixture, namely spurious-positive pairs. We believe that learning joint similarity helps to improve the performance when encoded features are distributed more evenly in the latent space.   We apply it as an orthogonal improvement for unsupervised contrastive representation learning, including current outstanding methods SimCLR \cite{chen2020simple}, MoCo \cite{he2020momentum}, BYOL \cite{grill2020bootstrap} and SimSiam \cite{chen2020exploring}. We evaluate our learned representations on many downstream benchmarks like linear classification on ImageNet-1k and PASCAL VOC 2007, object detection on MS COCO 2017 and VOC, etc. We obtain substantial gains with a large margin almost on all these tasks compared with prior arts.
\end{abstract}

\section{Introduction}

\begin{figure}[ht!]
	\centering
	\includegraphics[scale=0.2,width=0.5\linewidth]{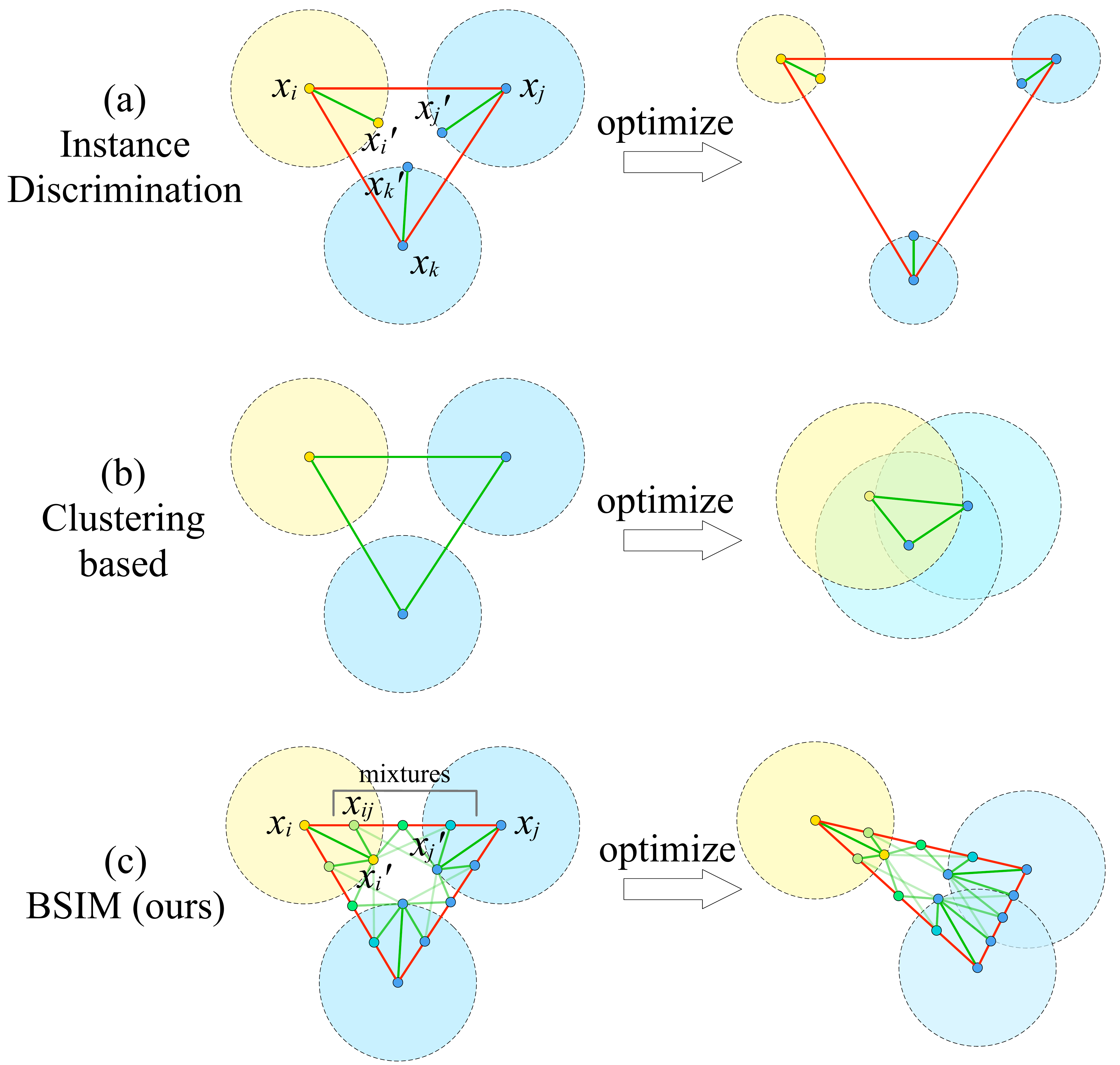}
	\vskip 0.1in
	\caption{A schematic view of three self-supervised paradigms. Note $x_i$, $x_j$ and $x_k$ are different instances. Green lines link the positive pairs while red for negative. Circles show decision boundaries (same color for the same class).  Instance discrimination narrows the boundary and pushes away all instances. Clustering-based methods might cluster wrong instances (\eg~yellow and blue) due to a shortcut defect. BSIM mixes instances (where hue indicates the ratio) to construct spurious positive pairs (\eg~($x_{ij}$, $x_{i}'$) and ($x_{ij}$, $x_j'$). BSIM encourages contrastive competition among instances, thus being better at learning interclass and intraclass representation. }
	\label{fig:intro}
	\vskip -0.25in
\end{figure}

Unsupervised representational learning is now on the very rim to take over supervised representation learning.
It is supposed to be a perfect solver for real-world scenarios full of unlabeled data.
Among them, self-supervised learning has drawn the most attention for its good data efficiency and generalizability.

Self-supervised learning typically involves a proxy task to learn discriminative representations from self-derived labels.
Among all manners of these proxy tasks \cite{noroozi2016unsupervised,gidaris2018unsupervised,pathak2016context,larsson2017colorization,donahue2016adversarial}, instance discrimination~\cite{wu2018unsupervised,liu2020self}, known as contrastive representation learning, has emerged as the most effective paradigm.
Its subsequent methods~\cite{zhuang2019local,he2020momentum,chen2020simple,grill2020bootstrap,tian2020makes}  have greatly reduced the gap between unsupervised and supervised learning.
Specifically, instance discrimination features a Single Instance Multi-view (SIM) paradigm to separate different instances. It seeks to narrow the distance among multiple views of the same instance (e.g. an image),  which are typically yielded from vanilla data augmentation policies like color jittering, cropping, resizing, applying Gaussian noise. 
Consequently, the invariance of the network is easily bounded by these limited augmentations.
Since different instances are continuously driven apart, SIM prevents itself from characterizing the relations among different instances from the same class, as opposed to supervised classification.

Meanwhile, clustering-based self-supervised methods \cite{caron2018deep,zhan2020online} alternate feature clustering with learning to capture similarities among different instances.
These methods avoid the intrinsic weakness of instance discrimination, but suffer from a so-called `shortcut' problem, i.e.,
when two instances are occasionally grouped into a cluster, their similarity will be further enhanced.
As a result, the training easily drifts into trivial solutions, \eg, merely grouping images in similar color or texture.

In view that instance discrimination pushes apart different instances indistinguishably as shown in Fig.~\ref{fig:intro}(a), and clustering-based methods are easily trapped in shortcut issues as shown in Fig.~\ref{fig:intro}(b),
we are motivated to explore a new paradigm to distinguish both intraclass and interclass instances.
In this work, we propose BSIM to learn better representations that capture high-level inter-image relations, which also potentially avoid the above-mentioned shortcut issue.
To make the minimal modification from previous works, BSIM shares similar pipelines to SimCLR \cite{chen2020simple}, MoCo \cite{he2020momentum}, BYOL \cite{grill2020bootstrap} and SimSiam \cite{chen2020exploring}, while focusing on a new way to construct positive pairs.

Specifically, as shown in Fig.~\ref{fig:intro}(c), BSIM first creates mixtures using CutMix \cite{yun2019cutmix} among instances by proportion, \eg, $x_{ij}$ that mixes $\lambda$ of $x_i$ and $(1-\lambda)$ of $x_j$, where $\lambda\in (0,1)$ obeys a Beta distribution.
Different from instance discrimination that constructs positive pairs between two views from the same image, BSIM makes use of mixed views $x_{ij}$ to create what we call \emph{spurious-positive} pairs $\left(x_{ij},x_i'\right)$ and $\left(x_{ij},x_j'\right)$. 
The optimization also proportionally takes $\lambda$ into account for computing the losses.
The interaction of spurious-positive pairs compete for an equilibrium state when grouping intra-instance and inter-instance views, modulated by the distribution of $\lambda$\footnote{It is worth noting that $\lambda\sim Bernoulli(0.5)$ degenerates the problem into instance discrimination exactly.}.
Meantime, negative pairs keep pushing away different instances.
As a result, BSIM encourages contrastive competition among instances, leaning towards exploring higher-level inter-image relations.
Since BSIM does not maintain dynamically changing pseudo labels as clustering-based methods, the shortcut issue is naturally avoided.
The contribution of this paper is twofold,
\begin{itemize}
	\item We propose a novel paradigm, namely BSIM, to encourage contrastive competition among instances for higher-level representation learning. Specifically, we generate \emph{spurious positive examples} using CutMix mixture, and we quantitatively score the distance between any image pairs by formulating a new contrastive loss . 
	\item BSIM is a general-purpose enhancement to existing methods that rely on instance discrimination (e.g. SimCLR, MoCo, BYOL, SimSiam). BSIM boosts performance for prior arts by clear margins and the gain from BSIM (e.g. BYOL-BSIM) is even comparable to the latest elaborately designed methods such as SimSiam \cite{chen2020exploring}.
	Moreover, it requires minimal modification to current self-supervised learning frameworks while adding neglectable cost. In general, BSIM-powered networks achieve state-of-the-art performance in a large body of standard benchmarks. 
\end{itemize}
\section{Related Work}

\noindent\textbf{Self-supervised learning based on contrastive loss.} 
Early methods focus on devising proxy tasks to either reconstruct the image after transformations \cite{larsson2017colorization,pathak2016context,zhang2017split}, or predict the configurations of applied transformation on a single image \cite{doersch2015unsupervised,dosovitskiy2014discriminative,noroozi2016unsupervised,gidaris2018unsupervised}.  Till recently contrastive loss based approaches \cite{he2020momentum,chen2020simple,grill2020bootstrap,Caron2020UnsupervisedLO} emerge as the mainstream paradigm, which features two components: the selection of positive or negative examples and the contrastive loss design. This routine leverages different augmented views of an image to construct positive pairs, while deeming other images as negative samples.  

Particularly, SimCLR \cite{chen2020simple} produces positive and negative pairs within a mini-batch of training data and chooses InfoNCE  \cite{oord2018representation} loss to train the feature extraction backbone. It requires a large batch-size to effectively balance the positive and negative ones. MoCo \cite{he2020momentum} makes use of a  feature queue to store negative samples, which  greatly reduces high memory cost in \cite{chen2020simple}. Moreover, it proposes a momentum network to boost the consistency of features. BYOL \cite{grill2020bootstrap} challenges the indispensability of negative examples and achieves impressive performance by only using positive ones.  A mean square error loss is applied to make sure that positive pairs can predict each other.  SimSiam \cite{chen2020exploring} utilizes stop-gradient as an alternative method to avoid mode collapse, simplifying the design compared to prior arts.

Besides, carefully designed augmentations to build positive pairs are proven to be critical for good performance \cite{chen2020simple,tian2020makes,chen2020improved,misra2020self,tian2019contrastive,asano2019critical,gontijo2020affinity,wang2020hypersphere}, because appropriate augmentations modulate the distribution of positive examples in the feature space.  SwAV \cite{Caron2020UnsupervisedLO} obtains the state-of-the-art unsupervised performance by using a mixture of views in different resolutions in place of two full-resolution ones. In the meantime, some researches study the role of hard negative examples \cite{Iscen_2018_CVPR,kalantidis2020hard,chuang2020debiased,Xie2020DelvingII,Wu2020OnMI}. However, all the above approaches try to push each image instance away from each other by regarding them as its negative samples. Is it possible to model the vicinity relation by measuring that distance quantitatively?  To our best knowledge, this problem is rarely studied in the field of self-supervised learning and BSIM is aimed to bridge this gap.

\noindent\textbf{Mixture as a regularization technique in supervised learning.}
Mixture of training samples like Mixup \cite{zhang2018mixup}, CutMix \cite{yun2019cutmix}, and Manifold Mixup \cite{verma2019manifold} has been proved to be a strong regularization for supervised learning, based on the principle of vicinal risk minimization \cite{chapelle2001vicinal}. It is designed to model vicinity relation across different classes other than vanilla data augmentation tricks that only considers the same class. CutMix \cite{yun2019cutmix} debates that Mixup introduces unnatural artifacts by mixing the whole image region while Cutout \cite{devries2017improved} might pay attention to less discriminative parts. CutMix claims to effectively localize the two classes instead. \cite{tokozume2018between} argues that mixing images is akin to mixing sounds \cite{tokozume2018learning} for CNNs, although not easily perceptible for humans. Other than deeming it as vanilla data augmentation that adds data variation, they consider it as an enlargement of Fisher's criterion \cite{fisher1936use}, \textit{i.e.}, the ratio of the between-class scatter to the within-class scatter, and it regularizes positional relationship among latent feature distributions. Furthermore, \cite{thulasidasan2019mixup} notices that label smoothing during mixup training has a calibration effect which regularizes over-confident predictions. 


\section{Methodology}

Different from the wide use of augmentation as a useful regularization  in supervised learning,  how to use it in unsupervised learning remains to be an open problem. Using the mixture solely as one of data augmentation techniques to produce positive pairs \emph{significantly} weakens the performance of MoCoV2 \cite{chen2020improved} (Table~\ref{tab:mix_strategy_drop}). This preferred regularization in supervised learning seems inherently incompatible with contrastive learning: drawing near multiple augmented views of an image while pushing away from the others, which we call Single Instance Multi-view (SIM) for simplicity. Instead, a mixture view shall be drawn close to both source images. This motivates us to scheme an alternative strategy for contrastive learning.  There are two basic and coupled problems to be answered: how to address the degradation and how to design feasible mixtures. 

\begin{table}[ht]
			\setlength{\tabcolsep}{1pt}
	\begin{center}
		\small
		\footnotesize
		\begin{tabular}{|l|c|c|c|}
			\hline
			Method &Epoch& SVM @VOC2007 & LC @ImageNet \\
			\hline\hline
			MoCoV2 \shortcite{he2020momentum} &200&83.81\% & 67.5\%\\
			MoCoV2+MixAug &200 & 80.10\% (-3.71\%$\downarrow$) & 64.6\% (-2.9\%$\downarrow$)\\ 
			\hline
		\end{tabular}
	\end{center}

	\caption{Regarding mixture as an extra data augmentation (MoCoV2+MixAug) weakens its performance severely. LC: linear classification on ImageNet.} 
	\vskip -0.1in
	\label{tab:mix_strategy_drop}
\end{table}

The central principle of contrastive learning is to encode semantically similar views (positive pairs) into latent representations that are close to each other while driving dissimilar ones (negative pairs) apart. 
A major question is how to effectively \emph{synthesize positive and negative pairs} given a dataset of i.i.d. samples as raised by \cite{tian2020makes}. Another question engages the \emph{design of contrastive loss}. Next, we discuss BSIM in details to address both.

\subsection{Spurious-Positive Views From Multiple Images}
 Given a set of images $\mathcal{D}$, two images $x_1,x_2$ sampled uniformly from $\mathcal{D}$, and two image augmentation distributions of  $\mathcal{T}'$ and $\mathcal{T}''$ (whether $\mathcal{T}'$ and $\mathcal{T}''$ are the same depends on different methods), we first generate a new example $x_{1,2}'$ by mixing $t'(x_1)$ and $t'(x_2)$, where $t'\sim \mathcal{T}'$. Specifically,  $x_{1,2}'$ borrows $\lambda$ region from $t'(x_1)$  and  remaining ($1-\lambda$) part from  $t'(x_2)$.  Thus, $x_{1,2}'$ has two spurious-positive examples, \ie, $t''(x_1)$ and $t''(x_2)$ , where $t''\sim \mathcal{T}''$.  These images are encoded by a neural network $f$ to extract high-level features, followed by a projection  head $g$  that maps the representation to a space ready to apply contrastive loss. The projection function is often implemented as a simple MLP network. For a better understanding, we give a schematic construction under SimCLR framework in Fig.~\ref{fig:bsim_simclr_framework} (supp.).
 
The distance between the newly mixed example and its parents is controlled by $\lambda$. BSIM uses a popular and handy option to generation $\lambda$ from Beta distributions,  \ie, $\lambda \sim \beta$($\alpha$, $\alpha$), where $\alpha$ is a hyper-parameter. It is evident that it would degrade to the single-instance multi-view case if $\lambda$ is always 0 or 1 when $\alpha \rightarrow 0$. That being said, BSIM is a generalization of SIM so that previous SIM methods reside within our larger framework.


\subsection{Loss Functions of BSIM}
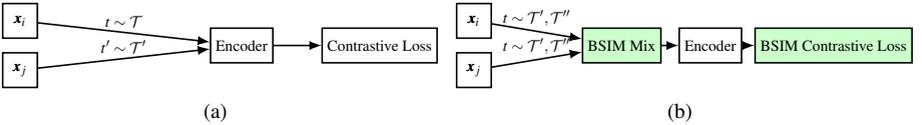
\begin{figure}[ht]
\small
\centering
\subfigure[]{
\begin{tikzpicture}[auto,node distance=0.65cm,
	semithick,scale=0.65, every node/.style={scale=0.65}]
\tikzstyle{var}=[draw, rectangle, minimum size=20pt]
\tikzstyle{arr}=[>=latex]
\node[var] at (-0.5,0.5) (xi) {$\bm x_i$};
\node[var] at (-0.5,-0.5) (xj) {$\bm x_j$};
\node[var] at (4,0) (en) {Encoder};
\node[var] at (6.8,0) (cl) {Contrastive Loss};
\path[->] 
 	(xi)  edge [arr] node[above,rotate=0] {$t \sim \mathcal{T}$} (en)
	(xj)  edge [arr] node[above,rotate=0] {$t' \sim \mathcal{T}'$} (en)
	(en)  edge [arr] node[above,rotate=0] {} (cl);
\end{tikzpicture}
}
\subfigure[]{
\begin{tikzpicture}[auto,node distance=0.65cm,
	semithick,scale=0.65, every node/.style={scale=0.65}]
\tikzstyle{var}=[draw, rectangle, minimum size=20pt]
\tikzstyle{arr}=[>=latex]

\node[var] at (-0.5,-1.5) (xib) {$\bm x_i$};
\node[var] at (-0.5,-2.5) (xjb) {$\bm x_j$};
\node[var, fill=green!20] at (2.5,-2) (bmix) {BSIM Mix};
\node[var] at (4.3,-2) (enb) {Encoder};
\node[var, fill=green!20] at (6.8,-2) (clb) {BSIM Contrastive Loss};
\path[->] 
 	(xib)  edge [arr] node[above,rotate=0] {$t \sim \mathcal{T}',\mathcal{T}''$} (bmix)
	(xjb)  edge [arr] node[above,rotate=0] {$t \sim \mathcal{T}',\mathcal{T}''$} (bmix)
	(bmix) edge [arr] node[above,rotate=0] {} (enb)
	(enb)  edge [arr] node[above,rotate=0] {} (clb);
\end{tikzpicture}
}
\vskip 0.1in
\caption{Our generic BSIM framework (b) serves as a plug-and-play adds-on for current contrastive learning paradigm (a). Note $\mathcal{T}$ and  $\mathcal{T}'$ are augmentation policy distributions. } 
\label{fig:bsim-framework}
\vskip -0.2in
\end{figure}

It's intuitive to change loss functions since spurious-positive examples are introduced. For example, it's unreasonable to assign  a new instance generated by half mixing two images (a dog and a cat, $\lambda=0.5$) to a dog or  cat. Human would easily tell this image is half cat and half dog. This means its projected feature in the high-dimensional latent space should be nearby a dog, as well as a cat, but much far from an orangutan.  This motivates us to design a particular loss for BSIM on top of the spurious-positive views, shown in Fig.~\ref{fig:bsim-framework}. Specifically, we adapt our method to four popular frameworks SimCLR \cite{chen2020simple}, MoCo \cite{he2020momentum}, BYOL \cite{grill2020bootstrap} and SimSiam \cite{chen2020exploring}. 
In order to make our paper more readable, we roughly follow the same naming conventions as these papers and list the symbol notations in Table \ref{tab:list-sym} (supp.).

\paragraph{SimCLR-BSIM.}
SimCLR uses a single augmentation distribution, i.e. $\mathcal{T}'$ and $\mathcal{T}''$ are identical herein.  The encoder network $f$ encodes $x_{1,2}' $ as $f(x_{1,2}')$.   Note $x_{1,2}'$ should show similarities with $x_1''$  as well as $x_2''$, which is measured  by the $\mathrm{sim}$ function in the projected $z$ space. We follow the definition in \cite{chen2020simple} for the similarity function as $\mathrm{sim}(\bm z_i, \bm z_j) = \bm z_i^\top \bm z_j / (\lVert\bm z_i\rVert \lVert\bm z_j\rVert)$. We use $\lambda$ to regularize these similarities and the matching loss can be formulated as,

\begin{equation}
\label{eq:loss}
\begin{split}
\ell'_{i}(\lambda)\!=\!-\lambda&\log \frac{e^{\mathrm{sim}(\bm z_{i,j}', \bm z_i'')/\tau}}{\sum_{k=1}^{N} [e^{\mathrm{sim}(\bm z_{i,j}', \bm z_k'')/\tau}\!+\!\one{} \cdot e^{\mathrm{sim}(\bm z_{i,j}', \bm z_{i,k}')/\tau}]} \\
\!-\!(1\!-\!\lambda)&\log \frac{e^{\mathrm{sim}(\bm z_{i,j}', \bm z_{j}'')/\tau}}{\sum_{k=1}^{N} [e^{\mathrm{sim}(\bm z_{i,j}', \bm z_k'')/\tau}\!+\!\one{}  \cdot e^{\mathrm{sim}(\bm z_{i,j}', \bm z_{i,k}')/\tau}]}. \\
\text{where\quad} \one{} &= \begin{cases}
1 & k \not\in \{i, j\} \\
0  & \text{otherwise}
\end{cases}
\end{split}
\end{equation}
Similarly, we can formulate $\ell_{i}''$ if we use $x_{1,2}''$ as the anchor. Hence, the NT-Xent  \cite{chen2020simple} loss is defined by the summation of each individual loss within the mini-batch data of size $N$ as,
\begin{equation}
\label{eq:loss-nt-xent}
L_{\texttt{NT-Xent}}(\lambda)=\frac{1}{2N}\sum_{k=1}^{N}\ell_{i}'(\lambda)+\ell_{i}''(\lambda),  \Hquad \lambda \sim \beta(\alpha,\alpha).
\end{equation}

SimCLR \cite{chen2020simple} has $2N$ positive pairs and $2N(N-1)$ negative ones in total at each iteration. Whereas, our method includes $4N$ spurious-positive pairs, i.e., $(x_{i,j}',x_{i}'')$, $(x_{i,j}',x_{j}'')$, $(x_{i,j}'',x_{i}')$, $(x_{i,j}'',x_{j}')$, and $2N(N-2)$ negative ones. The proposed method is depicted in Figure~\ref{fig:bsim_simclr_framework} (supp.).

\paragraph{MoCo-BSIM.}
  We produce the query $q$ of MoCo by forwarding the mixed image controlled by $\lambda$.  We illustrate the procedure in Fig.~\ref{fig:moco-bsim-framework} (supp.). 

\begin{equation}\label{eq:MoCO-BSIM}
	\small
\mathcal{L}_q =-\lambda \log \frac{\exp(q \cdot k^\lambda_{+}/\tau)}{\sum_{i=1}^{N}\exp(q\cdot k_{i}/\tau)}
	-(1-\lambda)\log \frac{\exp(q\cdot k^{1-\lambda}_{+}/\tau)}{\sum_{i=1}^{N}\exp(q\cdot k_{i}/\tau)}
\end{equation}
where $k^\lambda_+$ and $k^{1-\lambda}_+$ represent the corresponding key of images that produced the mixture respectively, and $k_i$ are the keys in the current queue. $\tau$ is the softmax temperature.

\paragraph{BYOL-BSIM.}
BYOL-BSIM generates two \augview s $x_1' \eqdef t'(x_1)$ and $x_1''\eqdef t''(x_1)$ from $x_1$ by applying respectively \imageaugmentation s $t'\sim \mathcal{T}'$ and $t''\sim \mathcal{T}''$.
Following the same procedure, we produce $x_2'$ and $x_2''$. Then we produce a new image $x'_{1,2}$ by $\lambda$-based mixture $x_1'$ and $x_2'$ through cutmix. The online network outputs  $y'_\netparams \eqdef f_\netparams(x'_{1,2})$ and the projection $z'_\netparams \eqdef g_\netparams(y')$. The target network yields two  $\ell_2$-normalized projections   $\bar{z}''_1$,  $\bar{z}''_2$    from $x''_1$ and $x''_2$. 

We sum up the MSE loss between the projection of the mixed image and its parents by the mixture coefficient $\lambda$. This process is shown in Fig.~\ref{fig:byol-bsim-lossv2} (supp.). Formally, the loss is:
\begin{equation}
\small
\begin{split}
\mathcal{L}'_{\netparams, \targetparams}=
-2[\lambda\frac{\langle q'_\netparams(z'_\netparams),  z''_{i,\targetparams} \rangle }{\big\|q'_\netparams(z'_\netparams)\big\|_2\cdot
	\big\|z''_{i,\targetparams}\big\|_2
} + (1-\lambda)\frac{\langle q'_\netparams(z'_\netparams),  z''_{j,\targetparams} \rangle }{\big\|q'_\netparams(z'_\netparams)\big\|_2\cdot
\big\|z''_{j,\targetparams}\big\|_2
}]
\end{split}
\label{eq:cosine-lossv0}
\end{equation}
Note $z''_{i,\targetparams}$ and $z''_{j,\targetparams}$ mean the projection of the representation of $x_i''$ and $x_j''$ generated by the target network.
We obtain $\mathcal{L}''_{\netparams, \targetparams}$ by using  $x_1''$ and $x_2''$ as the input of online network. Note that BYOL doesn't rely on negative samples. The normalized projection is on the sphere of a unit ball in the high dimensional space, see Fig.~\ref{fig:unit-ball} (supp.).

\paragraph{SimSiam-BSIM.} Since SimSiam utilizes the same loss as BYOL, we use exactly the same loss form as Eq~\ref{eq:cosine-lossv0} with scale 0.5 to match the loss in SimSiam \cite{chen2020exploring}.

\paragraph{WBSIM.} Alternatively, we offer BSIM as a general adds-on by adding a weighted BSIM (WBSIM) loss to the usual SIM losses, see Sec~\ref{sec:supp-wbsim} (supp.) for details.

\subsection{ Mixture Strategy Design}
CutMix and Mixup \cite{zhang2018mixup} are two popular  strategies of generating mixtures at image level. Whereas we don't utilize Mixup because it is less natural, even humans cannot easily tell the mixture coefficient $\lambda$ simply by checking the mixed image. We compare Mixup with CutMix via carefully controlled experiments (both with BSIM loss and the same hyper-parameter settings) under the framework of MoCoV2. Results from Table~\ref{tab: mix strategy strategy} disapprove of the use of Mixup in producing spurious-positive examples. The observation differs from supervised learning, where both boost the performance.

\begin{table}[ht]
	\begin{center}
		\footnotesize
		\begin{tabular}{|l|c|c|}
			\hline
			Method & SVM & SVM Low-Shot (96)  \\
			\hline\hline
			MoCoV2 \cite{he2020momentum} &83.81\% &82.01$\pm$0.13\%\\
			MoCoV2 (w/ Mixup, $\alpha$=1.0) & 82.50\% &80.54$\pm$0.26\% \\
			MoCoV2 (w/ Mixup, $\alpha$=0.5) & 82.80\% &80.58$\pm$0.31\% \\
			\textbf{MoCoV2-BSIM} (w/ CutMix, $\alpha$=1.0) & \textbf{84.55\%} & \textbf{82.65$\pm$0.34\%}\\
			\hline
		\end{tabular}
	\end{center}
	\caption{SVM evaluations on PASCAL VOC2007. Mixup deteriorates the performance of the baseline. } 
	\label{tab: mix strategy strategy}
	\vskip -0.1in
\end{table}

We also compare their performance using VOC object detection under the same metrics in Sec~\ref{sec:voc_od} (supp.).  The result is shown in  Table~\ref{tab:voc-od-ablation}. Mixup fails to improve the performance of its baseline without mixtures. In contrast, CutMix  can improve the detection performance. Therefore, we utilize CutMix as our default mixture strategy.

\begin{table}[ht]
	\begin{center}
		\footnotesize
		\begin{tabular}{|l|l|l|l|}
			\hline
			Method & AP$_{50}$ & AP$_{75}$ & AP \\
			\hline\hline
			No Mix& 82.4\%&63.6\%&57.0\%\\
			Mixup ($\alpha$=1.0)  &82.2\%(-0.2$\downarrow$) &63.4\%(-0.2$\downarrow$)  &56.9\%(-0.1$\downarrow$)\\
			\textbf{CutMix} ($\alpha$=1.0)&\textbf{82.7\%}(+0.3$\uparrow$) & \textbf{64.0\%}(+0.4$\uparrow$) &\textbf{57.3\%}(+0.3$\uparrow$)\\
			\hline
		\end{tabular}
	\end{center}
	\caption{Object detection results under the MoCoV2 framework on PASCAL VOC \texttt{trainval07+12}.}
	\label{tab:voc-od-ablation}
	\vskip -0.25in
\end{table}

\section{Experiments}
\paragraph{Setup.} We generally follow the compared methods. Details and cost are in Sec~\ref{sec:detailed_setting} (supp.). Object detection and instance segmentation experiments are given in Sec~\ref{sec:supp_od_is} (supp.).

\subsection{Evaluation on Linear Classification}

\paragraph{Linear SVM classification on VOC2007.}  The results are shown in Table~\ref{tab:svm}. In most cases, BSIM consistently boosts the baselines by about $1\%$ mAP. Particularly, BYOL-BSIM is boosted by a clear margin: \textbf{1.4\%} mAP. BYOL-BSIM300 outperforms the supervised pre-trained baseline with \textbf{0.4\%} higher mAP. BYOL-BSIM (200 epochs' training) is comparable to BYOL300 (300 epochs). Noticeably, WBSIM further boosts the performance. MoCoV2 benefits \textbf{1\%} mAP from BSIM, and an extra \textbf{0.6\%} higher mAP from WBSIM, indicating that BSIM is complementary to SIM.

\begin{table}[ht]
	\setlength{\tabcolsep}{0.8pt}
	\footnotesize
	\begin{center}
		\begin{tabular}{|l|H*{8}cc|}
			\hline
			\multirow{2}*{Method}  &Epoch& SVM & \multicolumn{8}{c|}{SVM Low-Shot (\%mAP)}\\
			& &\%mAP &1	&2&	4&	8	&16&	32&	64	&96 \\
			\hline\hline
			Supervised&-&87.2&53.0&63.6&73.7&78.8&81.8&83.8&85.2&86.0\\
			\hline
			\hline
			SimCLR \shortcite{chen2020simple}&200&79.0&32.5 &	40.8&	50.4	&59.1&65.5&70.1&73.6&75.4\\
			\rowcolor{blue!10} SimCLR-BSIM&200 &80.0 & 33.9& 44.7&50.9& 60.5&67.8&72.0&75.4&77.2\\
			\hline
			MoCo \shortcite{he2020momentum} &200& 79.2&30.0&	37.7&	47.6	&58.8 &	66.0	&70.6	&74.6 &	76.1 \\
			MoCoV2 \shortcite{chen2020improved}&200& 83.8& 43.7 &	55.2& 63.2&71.5&75.4&79.1&81.2	&82.0 \\
			\rowcolor{blue!10} \textbf{MoCoV2-BSIM} &200& 84.8&  \textbf{50.0}& 53.9& \textbf{65.3}&72.4&76.3&79.3&81.7&82.8 \\
			\rowcolor{blue!10} \textbf{MoCoV2-WBSIM}&200 & 85.4&46.5&56.9&64.6&\textbf{74.7}&78.2&80.6& 82.8& 83.7\\
			\hline
			BYOL \shortcite{grill2020bootstrap}&200& 85.1& 44.5&52.1&62.9&70.9	&76.2&79.5&81.9&83.1\\
			\rowcolor{blue!10} \textbf{BYOL-BSIM}&200 & \textbf{86.5} & 42.6&\textbf{55.9}&64.6&72.7&\textbf{78.8}&\textbf{81.9}&\textbf{83.6}&\textbf{84.6}\\
			\hline
			BYOL300 \shortcite{grill2020bootstrap} &300&86.6& 42.5 & 56.1& 64.7 & 73.0&77.7& 82.2& 83.7& 84.7 \\
			\rowcolor{blue!10} \textbf{BYOL-BSIM300} &300&87.6& \textbf{45.7} &54.5&66.4&75.0&79.8&83.2&\textbf{85.2}&86.0\\
			\rowcolor{blue!10 }\textbf{BYOL-WBSIM300}&300&\textbf{87.7}&44.1&\textbf{60.7}& \textbf{68.1}&\textbf{76.0}&\textbf{81.0}&\textbf{83.6}&\textbf{85.2}&\textbf{86.3} \\
			\hline
			SwAV \shortcite{caron2020unsupervised}$^\star$ & 400 & 85.4 &-&-&-&-&-&-&-&- \\\hline
		\end{tabular}
	\end{center}
	\caption{ResNet-50 linear SVMs mAP on VOC07 \protect\cite{everingham2010pascal} classification using two 224$\times$ 224 views. BYOL variants with ``300'' are trained for 300 epochs as \protect\cite{grill2020bootstrap}. $^\star$: SwAV is trained for 400 epochs. } 
	\label{tab:svm}
	\vskip -0.2in
\end{table} 
%

 \paragraph{Low-shot classification on VOC2007.}  The results are shown in Table~\ref{tab:svm}. BSIM helps all baselines to achieve better performance by substantial margins. It's interesting to see that BYOL-BSIM300 gradually bridge its gap from the supervised baseline. When the number of the training set is more than 64, it's comparable to the supervised version.

\paragraph{Linear classification on ImageNet.} 
The results are shown in Table~\ref{tab:linear-imagenet}, where the performance of the competing methods are extracted from \cite{zhan2020open}. 

\begin{table}[ht]
	\footnotesize
	\begin{center}
		\setlength{\tabcolsep}{2pt}
		\begin{tabular}{|l|c|c|HHHHHHl|l|l|}
			\hline
			Method &Epoch &Backbone & &&&& &&Top-1 Accuracy\\
			\hline\hline
			InfoMin Aug \shortcite{tian2020makes}&200&R50&-&-&-&-&70.1&-&70.1\\
			MoCo \shortcite{he2020momentum} & 200&R50&15.3&33.1&44.7&57.3&60.6& 61.0 & 61.0\\
			\hline
			SimCLR\shortcite{chen2020simple}&200&R50&17.1&31.4&41.4&54.4&61.6&60.1& 61.6\\
			\rowcolor{blue!10}\textbf{SimCLR-BSIM} & 200&R50& \textbf{18.0} & 32.5 & 42.7& 55.3& 62.3 (+0.7$\uparrow$)& 60.7&62.3 (+0.7$\uparrow$)\\
			
			\hline
			MoCoV2 \shortcite{chen2020improved}&200 &R50&14.7&32.8&45.0	&61.6&66.7&67.5&67.5\\
			\rowcolor{blue!10}MoCoV2-BSIM & 200&R50&15.7& 34.2 &46.8 & 63.1 & 67.6 &  68.0 (+0.5$\uparrow$)&68.0 (+0.5$\uparrow$)\\
			\rowcolor{blue!10}MoCoV2-WBSIM & 200 &R50& 16.0&35.0  &48.1&64.7 &68.2 &68.4 (+0.9$\uparrow$) & 68.4 (+0.9$\uparrow$)\\
			\hline
			BYOL \shortcite{grill2020bootstrap}& 200&R50&16.7&34.2&46.6&60.8&69.1	  &67.1 & 69.1 \\
			\rowcolor{blue!10}BYOL-BSIM & 200&R50& 17.5 & 35.1&47.4&62.0 &69.8 (+0.7$\uparrow$) &67.9  &69.8 (+0.7$\uparrow$)\\
			\hline
			BYOL \shortcite{grill2020bootstrap}$^\dagger$ & 300 &R50&14.1&34.4&47.2&63.1&72.3&70.3& 72.3\\
			\rowcolor{blue!10}\textbf{BYOL-BSIM} & 300 &R50&16.4& \textbf{35.3}&48.5&65.1&  72.7 (+0.4$\uparrow$)& 70.7&  72.7 (+0.4$\uparrow$)\\
			\rowcolor{blue!10}\textbf{BYOL-WBSIM}&300&R50 &15.4& \textbf{35.3}&\textbf{48.7}&\textbf{65.7}&   \textbf{73.0} (\textbf{+0.7$\uparrow$})&\textbf{71.1}& \textbf{73.0} (\textbf{+0.7$\uparrow$})\\
			\hline
			SimSiam \shortcite{chen2020exploring} & 200&R50 & -& -& -& -&70.0 & - & 70.0 \\
		    \rowcolor{blue!10}SimSiam-BSIM \shortcite{chen2020exploring} & 200 &R50& -& -& -& -&70.4(+0.4$\uparrow$)& - &70.4(+0.4$\uparrow$)\\
			\rowcolor{blue!10}SimSiam-WBSIM \shortcite{chen2020exploring} & 200 &R50& -& -& -& -&70.8(+0.8$\uparrow$)& - &70.8(+0.8$\uparrow$) \\
			\hline
			SwAV \shortcite{caron2020unsupervised}&200&R50 & -&-& -&-&69.1&- &69.1\\
			SwAV \shortcite{caron2020unsupervised}&400 &R50& -&-& -&-&70.7&- &70.7\\
			\hline
		\end{tabular}
	\end{center}
	\caption{Linear classification on ImageNet (top-1 center-crop accuracy on the validation set). All models are trained with two 224$\times$224 views.  $^\dagger$: reproduced. SwAV result is from SimSiam \protect\cite{chen2020exploring}. } 
	\label{tab:linear-imagenet}
	\vskip -0.3in
\end{table}

\subsection{Evaluation on Semi-supervised Classification}

Results are shown in Table~\ref{tab:semi-cls}. BSIM improves the baselines by significant margins, especially when the amount of available labels is small. MoCoV2-BSIM obtains $44.3\%$ top-1 accuracy, which is $5.2\%$ higher than MoCoV2. Although BYOL-BSIM is only trained for 200 epochs, it achieves comparable results as BYOL1000. WBSIM300 can further boost the performance to the state-of-the-art 57.2\%. Specifically, it obtains 57.2$\%$ top-1 accuracy using 1$\%$ labeled data, which is about 4$\%$ higher than BYOL1000.  When we collect more data ($10\%$),  BYOL-WBSIM  outperforms BYOL1000 with about $2\%$. Combining SIM and BSIM seems to learn better representations.

\begin{table}[ht]
	\footnotesize
	\begin{center}
		\setlength{\tabcolsep}{0.8pt}
		\begin{tabular}{|l|c|ll|ll|}
			\hline
			\multirow{2}*{Method} &\multirow{2}*{Epoch}&\multicolumn{2}{c|}{1\% labelled data} & \multicolumn{2}{c|}{10\% labelled data} \\
			&&top-1($\%$) &top-5($\%$)&top-1($\%$)    &top-5($\%$)\\
			\hline\hline
			Supervised &-&68.7&88.9&74.5&92.2\\
			\hline			
			SimCLR \shortcite{chen2020simple}&200&36.1	&64.5&58.5&82.6 \\
			\rowcolor{blue!10}SimCLR-BSIM &200 &38.2\inc{2.1} & 67.5\inc{3.0}&61.2\inc{2.7}& 84.5\inc{2.9}\\

			\hline
			MoCo \shortcite{he2020momentum} &200& 33.2	&61.3& 60.1&84.0\\
			MoCoV2 \shortcite{chen2020improved}&200&39.1	&68.3&61.8&85.1\\
			\rowcolor{blue!10}MoCoV2-BSIM &200&40.8\inc{1.7}&70.3\inc{2.0}&62.6\inc{0.8}&85.8\inc{0.7}\\
			\rowcolor{blue!10}MoCoV2-WBSIM &200&44.3\inc{5.2}&72.9\inc{4.6} &63.9\inc{2.1}&86.6\inc{1.5}\\
			\hline
			BYOL \shortcite{grill2020bootstrap}& 200&49.4&76.8&65.9&	87.8 \\
			\rowcolor{blue!10}\textbf{BYOL-BSIM} &200& \textbf{53.0}\inc{3.6} & \textbf{79.9}\inc{3.1} & \textbf{68.2} \inc{2.3}& \textbf{89.0} \inc{2.2}\\
			\hline
			SwAV \shortcite{caron2020unsupervised} &800& 53.9 & 78.5 & 70.2 & 89.9 \\
			SimCLR \shortcite{chen2020simple}&1000 &48.3&75.5&65.6&87.8 \\
			BYOL \shortcite{grill2020bootstrap} &1000 &53.2&78.4&68.8&89.0\\
			\rowcolor{blue!10}\textbf{BYOL-WBSIM} &300& \textbf{57.2}&\textbf{81.8} & \textbf{70.7}&\textbf{90.5} \\
			\hline
		\end{tabular}
	\end{center}
	\caption{Semi-supervised classification on ImageNet. We report center-crop accuracy on the val set.} 
	\label{tab:semi-cls}
	\vskip -0.2in
\end{table}



\section{Ablation and Discussions}\label{sec:discuss}

\paragraph{Sensitivity on $\alpha$.}
We further analyze the performance sensitivity of $\alpha$. Regarding the intensive resource cost, we report the SVM and low-shot SVM results in Table~\ref{tab: sensitivity_alpha} using MoCo-V2. We keep the same pre-training setting. The  distribution from group  $\alpha = 0.75$ performs best.
The performance keeps stable when $\alpha > 0.5$. However, it drops severely once $\alpha\rightarrow 0$ when it degenerates to MoCo-V2.

\begin{table}[ht]
	\begin{center}
		\footnotesize
		\setlength{\tabcolsep}{3pt}
		\begin{tabular}{|l|c|c|c|}
			\hline
			Method & $\alpha$ & SVM & SVM Low-Shot (96)  \\
			\hline\hline
			MoCoV2-BSIM &1.0 & 84.55&82.65$\pm$0.34 \\
			\textbf{MoCoV2-BSIM} &0.75 & \textbf{84.56}  &\textbf{82.67$\pm$0.26} \\
			MoCoV2-BSIM &0.5 & 84.23&82.50$\pm$0.29\\
			MoCoV2-BSIM &0.25 & 84.02 & 82.18$\pm$0.31\\			
			\hline
		\end{tabular}
	\end{center}
	\caption{Performance sensitivity on $\alpha$ using MoCoV2-BSIM. The classification result is averaged across 5 independent experiments. When $\alpha < 0.01$, MoCoV2-BSIM can be regarded as MoCoV2 approximately which achieves 83.8\% mAP on SVM.} 
	\label{tab: sensitivity_alpha}
	\vskip -0.3in
\end{table}

\paragraph{Why does mixture as data augmentation fail?}
As mentioned in Table~\ref{tab: mix strategy strategy}, simply adopting mixture methods as a data augmentation option severely degrades the performance.
Regarding the mixed image as the same instance as the original one forces the network to expand the decision boundaries blindly.
Consequently, the network might be trapped in shortcut solutions to group images in different classes indiscriminately.

\paragraph{Why  BSIM improves discrimination?}
A mixture is close to the decision boundaries in instance discrimination task where neural networks are normally less certain about.
In instance discrimination, the decision boundaries keep being separated, leaving some area crowded while others sparse, which is unfavorable for learning high-level inter-image relations.
In BSIM, when learning $\lambda$-balanced similarities between competing spurious-positive pairs, the network encourages contrastive competition among instances to occupy the area near decision boundaries (Fig.~\ref{fig:bsim-intra} supp.). We sample three times to demonstrate that BSIM in general gives cleaner inter-class representation, while SIM has a more intertwined one. This proves that learning the distance to the mixture helps improve classification by generating a better latent representation. 
In this way, we hypothesize that the network has to encode the latent representations more accurately like a ruler.
As a result, the features have to scatter more evenly, especially for between-class areas that are harder to predict, which is shown in  Fig.~\ref{fig:mixup-32} (supp.). We hypothesize this is a major factor that BSIM offers stronger discrimination capability. 
In Sec~\ref{sec:supp_dicuss} (supp.), we give a comparison with other mixture-based approaches and draw a latent representation via TSNE to manifest the working mechanism of BSIM.


\section{Conclusion}
In this paper, we propose BSIM, a novel self-supervised representation learning approach beyond the current instance discrimination paradigm.
It makes minimal modification to the existing instance discrimination methods such as SimCLR, MoCo, BOYL, and SimSiam, while significantly improving the performance on many downstream tasks.
We justify the superiority of BSIM via analyzing the optimization behaviors when combined with different paradigms, which provides a new perspective in the field of contrastive representation learning.
%
Being a simple and lightweight plugin, it substantially enhances the SSL performance. 

\bibliography{../icml22/egbib.bib}

\newpage

\appendix

\section{Index of Symbols}
To facilitate readability, we give a complete list of notations in Table~\ref{tab:list-sym}.
\begin{table}[ht]
	\begin{center}
		\begin{tabular}{|c|l|}
			\hline
			Symbol & Definition  \\
			\hline\hline
			$x_i$, $x_j$   & input sample \\ 
			$x_i'$, $x_i''$ & augmented sample (view) \\ 
			$\mathcal{T}'$, $\mathcal{T}''$      & augmentation distribution \\
			$h_i'$, $h_i''$ & feature representation \\ 
			$z_i'$, $z_i''$ & representation mapped to $z$-space  \\ 
			$\beta(\alpha,\alpha)$          & beta distribution \\
			$\lambda$     & sampled variable from $\beta$  \\
			$\tau$ & softmax temperature \\
			$q$ & query in the MoCo framework\\
			$k_{+}^{\lambda}, k_{+}^{1-\lambda}$ & key of the mixed samples \\
			$k_i$ & key in the current queue \\
			$f(\cdot)$, $f_{\theta}$ & encoder network \\
			$g(\cdot)$, $g_{\theta}$ & projection head \\
			$q_{\theta}$ & predictor \\
			$\ell_{i}'$ & contrastive loss with $x_{i,j}$ as an anchor \\
			$\mathcal{L}_{NT-Xent}(\lambda)$ & loss of SimCLR-BSIM \\
			$\mathcal{L}_{q}$ & loss of MoCo-BSIM \\
			$\mathcal{L}_{\theta, \xi}'$ & loss of BYOL-BSIM \\
			$\mathcal{L}_{WBSIM}$ & loss of weighted BSIM \\
			\hline
		\end{tabular}
	\end{center}
	\caption{List of symbols used throughout the paper.}
	\label{tab:list-sym}
\end{table}

\section{BSIM as a General Adds-on Approach}\label{sec:supp-wbsim}

Apart from the discussed integration with SimCLR, MoCo and BYOL, we can also simply treat BSIM as an adds-on to SIM-based methods by a weighted summation of loss functions,
\begin{equation}
\mathcal{L}_{WBSIM}= w_1*\mathcal{L}_{BSIM} + w_2 * \mathcal{L}_{SIM},
\end{equation}
where $w_1, w_2 \in \left(0, 1\right)$.
We refer this approach as weighted-BSIM (\textbf{WBSIM}). When $w_1=0,w_2=1$, it is the conventional single instance multi-view approach. When $w_1=1, w_2=0$, it is BSIM. We set $w_1=w_2=0.5$ throughout the paper to benefit from both SIM and BSIM.

Feature-level mixture is utilized as a regularization to perform hard example mining \cite{kalantidis2020hard}, which can boost discrimination. Other than using it as an extra augmentation, we focus on image-level mixture to define the spurious-positive examples and quantify how close two images are.

\section{Experiment Details}\label{sec:detailed_setting}

\subsection{Self-supervised Pre-training}

In self-supervised pre-training, we generally follow the default settings of the competing methods for fair comparisons. We freeze the weights of ResNet50. Unless otherwise specified, all methods are trained for 200 epochs on the ImageNet dataset.

\paragraph{SimCLR-BSIM.} We use the same set of data augmentations as \cite{chen2020simple}, \ie, random cropping, resizing, flipping, color distortions, and Gaussian blur. The projection head is a 2-layer MLP that projects features into 128-dimensional latent space. We use the modified NT-Xent loss as in Equation~\ref{eq:loss-nt-xent} and optimize with LARS \cite{you2017scaling} with the weight decay 1e-6 and the momentum 0.9. We reduce the batch size to 256, and learning rate to 0.3 with linear warmup for first 10 epochs and a cosine decay schedule without restart. 

\paragraph{MoCo-BSIM.} We follow \cite{he2020momentum}  for MoCo experiments. We first train ResNet-50 with an initial learning rate 0.03 for 200 epochs on ImageNet (about 53 hours on 8 GPUs) with a batch size of 256 using  SGD with weight decay 1e-4 and momentum 0.9. For downstream tasks, the model is finetuned with BNs enabled and synchronized across GPUs. For MoCoV1, we utilize a linear neck with 128 output channels and a  $\tau$ of 0.07. As for MoCoV2, we use two FC layers (2048, 2048, 128) to perform projections and  a temperature coefficient  of 0.2.

\paragraph{BYOL-BSIM.} Data augmentation is the same as \cite{grill2020bootstrap}.  We follow \cite{grill2020bootstrap} for the default hyper-parameters. Note that \cite{grill2020bootstrap} states they prefer 300 epochs to make comparisons, we also conform to it to be consistent. Since many methods report their performance on 200 epochs, we also add an extra setting of 200 epochs to make fair comparison. To differentiate these two versions, we use 200 by default and name the BYOL300 for the former. Specifically, we also optimize LARS \cite{you2017scaling} with weight decay $1.5\cdot10^6$. We set the initial learning rate 3.2 and use a batch size of 4096. The target network has an exponential moving average parameter $\tau=0.996$ and increased to 1. 

\paragraph{SimSiam-BSIM.}  We use the same setting as \cite{chen2020exploring}, which is similar to \cite{he2020momentum}. Note that the weight decay of 0.0001 is applied for all parameter layers, including batch normalization and bias.

\subsection{Downstream classification}

\paragraph{Linear classification on ImageNet for BYOL.} 

BYOL \cite{grill2020bootstrap} adopts a quite different setting.  To reproduce the baseline results, we train it for 90 epochs using the SGD optimizer with 0.9 momentum.  Besides, we use L2 regularization with 0.0001 and a batch size of 256.   The initial learning rate is 0.01 and scheduled by the 0.1 $\times$ at epoch 30 and 60. 

\paragraph{Linear SVM classification on VOC2007.} Following \cite{owens2016ambient,goyal2019scaling,zhan2020online}, we use the \texttt{res4} block (notation from \cite{girshick2018detectron})  of ResNet-50 as the fixed feature representations and train SVMs \cite{boser1992training} for classification using LIBLINEAR package \cite{fan2008liblinear}.  We train on the \texttt{trainval} split of the VOC2007 dataset \cite{everingham2010pascal} and report the mean Average Precision (mAP) on the \texttt{test} split by 3 independent experiments. All methods are evaluated using the same hyper-parameters as in \cite{goyal2019scaling}.  

\paragraph{Linear classification on ImageNet.} We follow the linear classification protocol in \cite{he2020momentum} where a linear classifier is appended to frozen features for supervised training. As for MoCo, we train 100 epochs using SGD with a batch size of 256. The learning rate is initialized as 30  and scheduled by 0.1 $\times$ at epoch 30 and 60.  

\paragraph{Low-shot classification on VOC2007.} We evaluate the low-shot performance when each category contains much fewer images. Following \cite{wang2016learning,goyal2019scaling,zhan2020online}, we use seven settings (N=1, 2, 4, 8, 16, 32, 64 and 96 positive samples), train linear SVMs on the low-shot splits, and report the test results across 3 independent experiments.

\paragraph{Evaluation on Semi-supervised Classification}

For semi-supervised training, we use the same split 1\% and 10\% amount of labeled ImageNet images as done in  \cite{zhai2019s4l,chen2020simple}. We follow \cite{chen2020simple,henaff2019data,kornblith2019better,zhai2019large} to finetune ResNet50's backbone on the labeled data. We train 20 epochs using SGD optimizer  (0.9 momentum) with a batch size of 256.  The learning rate is initialized as 0.01 and decayed by 0.2$\times$ at epoch 12 and 16.

\subsection{Object Detection and Instance Segmentation}\label{sec:supp_od_is}

\paragraph{Evaluation on PASCAL VOC Object Detection}\label{sec:voc_od}
Following the evaluation protocol by \cite{he2020momentum} where ResNet50-C4 (\ie, using extracted features of the 4-th stage) is used as the backbone and Faster-RCNN \cite{ren2015faster} as the detector, we benchmark our method for the object detection task on the VOC07 \cite{everingham2010pascal} \texttt{test} set. All models are finetuned on the \texttt{trainval} of VOC07+12 dataset for 24k iterations. We use Detectron2 \cite{wu2019detectron2}  like MoCo did. Results are reported in Table~\ref{tab:voc-od}, which are mean scores across five trials as \cite{chen2020improved} using the COCO suite of metrics \cite{lin2014microsoft}. Combined with BSIM, BYOL achieves 1.4$\%$ higher AP and 0.8$\%$ higher AP$_{50}$.

\begin{table}[ht]
	\footnotesize
	\begin{center}
		\begin{tabular}{|l|c|c|c|c|}
			\hline
			Method & Epoch&  AP$_{50}$ & AP$_{75}$ & AP \\
			\hline\hline
			supervised & -&81.3&58.8 & 53.5\\
			\hline
			SimCLR \shortcite{chen2020simple}&200&79.4&55.6& 51.5\\
			\rowcolor{blue!10}SimCLR-BSIM & 200& 79.8&56.0&51.8\\
			\rowcolor{blue!10}SimCLR-WBSIM &200& 80.0&56.2&51.9\\
			\hline
			MoCo \shortcite{he2020momentum} & 200&81.5&62.6& 55.9 \\
			MoCoV2 \shortcite{chen2020improved}&200& 82.4&63.6&57.0\\
			MoCoV2 \shortcite{chen2020improved}&200& 82.5&64.0&57.4\\
			\rowcolor{blue!10}MoCoV2-BSIM &200&82.7&64.0&  57.3\\
			\rowcolor{blue!10} \textbf{MoCoV2-WBSIM} &200&\textbf{83.0}&\textbf{64.2}&\textbf{57.5}\\
			\hline
			SimSiam, base \shortcite{chen2020exploring} &200&82.0& 62.8 & 56.4\\
			SimSiam, optimal \shortcite{chen2020exploring} &200&82.4& 63.7 & 57.0\\
			\rowcolor{blue!10}SimSiam-BSIM & 200& 82.8 & 64.0 & 57.3 \\
			\rowcolor{blue!10}\textbf{SimSiam-WBSIM}&200 & \textbf{83.0} & \textbf{64.2} & 57.4 \\
			\hline
			BYOL \shortcite{grill2020bootstrap}&200& 81.0& 56.5 & 51.9\\
			\rowcolor{blue!10}BYOL-BSIM &200& 81.8 & 58.4 & 53.3 \\
			\rowcolor{blue!10}BYOL-WBSIM &200& 82.0 & 58.5 & 53.5 \\\hline
			SwAV \shortcite{caron2020unsupervised}&800 & 82.6&62.7&56.1\\
			\hline
		\end{tabular}
	\end{center}
	\caption{Detection results on PASCAL VOC \texttt{trainval07+12}, which are reported across 5 trials. To make fair comparisons, the backbone is pre-trained for 200 epochs. MoCoV2-WBSIM trained 200 epochs surpasses  MoCoV2 and SwAV trained for 800 epochs.}
	\label{tab:voc-od}
\end{table}

\paragraph{Evaluation on  COCO Objection Detection and Instance Segmentation}
We also follow the evaluation protocol by \cite{he2020momentum} for the object detection and instance segmentation task on COCO2017  \cite{lin2014microsoft}. Specifically, we use the ResNet50-C4 Mask R-CNN framework \cite{he2017mask} and follow 2$\times$ schedule \cite{girshick2018detectron} as \cite{he2020momentum} since this setting can make fairer evaluations.   All models are fine-tuned on the \texttt{train2017} set and evaluated on \texttt{val2017}. We report the bounding box AP and mask AP on COCO in Table~\ref{tab:coco-od}. 

\begin{table}[t]
	\begin{center}
		\setlength{\tabcolsep}{1pt}
		\footnotesize
		\begin{tabular}{|l|c|c|c|c|c|c|c|c|}
			\hline
			Method & Epoch& AP$_{50}^{b}$ & AP$_{75}^{b}$ & AP$^{b}$ &  AP$_{50}^{m}$ & AP$_{75}^{m}$ &AP$^{m}$ \\
			\hline\hline
			Supervised& -&59.9&43.1&40.0&56.5&36.9&34.7\\
			\hline
			SimCLR \shortcite{chen2020simple}&200&59.1&42.9&	39.6&55.9&37.1&34.6\\
			\rowcolor{blue!10}SimCLR-BSIM &200&59.3& 43.1& 39.8 & 56.2 &37.4&34.8 \\
			\rowcolor{blue!10}SimCLR-WBSIM &200&59.5& 43.2& 40.0 & 56.4 &37.5&34.9 \\
			MoCo \shortcite{he2020momentum} &200& 60.5&44.1&40.7	&57.3&	37.6&35.4 \\
			MoCoV2 \shortcite{chen2020improved}&200& 60.1	&44.0&40.6&56.9&38.0&35.3\\
			\rowcolor{blue!10}MoCoV2-BSIM &200&60.3& 44.2&40.9 &57.0 &38.2 & 35.4\\
			\rowcolor{blue!10}\textbf{MoCoV2-WBSIM} &200&60.4& \textbf{44.4}& \textbf{41.1} & \textbf{57.2} & \textbf{38.3} & \textbf{35.5}\\
			BYOL \shortcite{grill2020bootstrap}&200&60.5&43.9&	40.3&56.8&37.3&35.1\\
			\rowcolor{blue!10}BYOL-BSIM &200& 60.8 &44.2& 40.7& 57.0& 37.5& 35.3\\
			\rowcolor{blue!10}\textbf{BYOL-WBSIM} &200& \textbf{61.0} &44.3& 40.9& \textbf{57.2}& 37.6& \textbf{35.5}\\
			\hline
			SwAV \shortcite{caron2020unsupervised} &800&59.8&42.0&39.1&56.2&36.1&34.2\\
			\hline
		\end{tabular}
	\end{center}
	\caption{Object detection and instance segmentation fine-tuned results on COCO2017 dataset using 2$\times$ schedule.  }
	\label{tab:coco-od}
\end{table}

\subsection{Training and Memory Cost} 

All the experiments are done on Tesla V100  with 8 GPUs. We use a batch size of 2048  for the BYOL experiment and accumulate gradients to simulate a batch size of 4096. MoCo-BSIM adds no extra memory cost to MoCo where we simply replace query samples with mixed ones. Whereas the WBSIM version has to maintain the originally augmented query samples to compute MoCo's default loss. See details in Table~\ref{tab:moco-mem-cost}.

\begin{table}[t]
	\begin{center}
		\begin{tabular}{|l|c|c|c|}
			\hline
			& MoCo & MoCo-BSIM& MoCo-WBSIM \\
			\hline\hline
			Memory(G)&5.5&5.5&8.2\\
			Cost (Hour)&53&53&65\\
			\hline
		\end{tabular}
	\end{center}
	\caption{Memory cost and training cost tested using a batch size of 256 across 8 GPUS. The training cost is calculated based on 200 epochs. }
	\label{tab:moco-mem-cost}
	\vskip 0.1in
\end{table}

\begin{table}[ht]
	\footnotesize
	\setlength{\tabcolsep}{2pt}
	\begin{center}
		\begin{tabular}{|l|c|c|c|}
			\hline
			Method & Batch Size & Memory (G) & GPU Days \\
			\hline
			MoCo &256 &44 &17.7\\
			\rowcolor{blue!10}MoCo-BSIM &256 &44 &17.7\\
			\rowcolor{blue!10}MoCo-WBSIM & 256 & 65.6 & 21.7 \\\hline
			BYOL & 4096 & 216 & 16 \\
			\rowcolor{blue!10}BYOL-BSIM & 4096 & 216 & 16 \\
			\rowcolor{blue!10}BYOL-WBSIM & 1024 & 200 & 28 \\
			SwAV & 4096 & 819 & 33.3\\
			\hline
		\end{tabular}
	\end{center}
	\caption{GPU resources cost. SwAV is tested on 64 V100-16G GPUs, others on 8 V100-32G GPUs. The training cost is calculated based on 200 epochs.}
	\label{tab:moco-mem-cost}
	\vskip -0.2in
\end{table}

\paragraph{The sampling process.} For a mini-batch of samples with size $N$, theoretically, we can sample $\lambda$ for $N$ times to enrich the information. Consequently, we can construct the loss by using $N$ different weighted items. However, this process can hardly be implemented efficiently in the PyTorch framework. Instead, we make use of the 8 GPU workers and set different seeds at the beginning of training. This approach is quite efficient and possesses rich mixtures. 

We also compare the performance of various methods given longer training epochs. The results are listed in Table~\ref{tab:lc_longer_epoch}.  Compared with MoCoV2, MocoV2-WBSIM can further improve $0.3\%$ top-1 accuracy on ImageNet validation dateset. Moreover, it can boost about 1 $AP$ on VOC detection task.

\begin{table}[ht]
	\begin{center}
	\setlength{\tabcolsep}{2pt}
	\footnotesize
		\begin{tabular}{|l|*{3}{c}|ccc|}
			\hline
			Method & Epoch & \multicolumn{2}{c|}{ImageNet Acc}&\multicolumn{3}{c|}{VOC Detection}\\
			&&Top-1 & Top-5&AP$_{50}$&AP$_{75}$&AP\\
			\hline\hline
			MoCoV2 \shortcite{chen2020improved}&800&71.1&- &82.5&64.0&57.4\\
			\rowcolor{blue!10}MoCoV2-WBSIM &800&71.4&90.4& 83.4& 65.0& 58.3 \\
			\hline
		\end{tabular}
	\end{center}
	\caption{Linear evaluation on ImageNet and  object detection on VOC  using ResNet50.}
	\label{tab:lc_longer_epoch}
\end{table}




\section{More Discussions}\label{sec:supp_dicuss}

\paragraph{Comparison with mixture-based approaches.}
Shen et al. \cite{shen2020rethinking} propose a somewhat complicated iterative mixture strategy exploiting Mixup \cite{zhang2018mixup} and CutMix \cite{yun2019cutmix} to generate a weighted mixture of samples. The mixture can be considered a weakened version of the original images which is harder to recognize, hence rendering flattened predictions. As suggested from the label-smoothing perspective, it is meant to suppress incorrect response on hard negative samples. However, image mixtures are used as-is, i.e., it learns the mixture-to-mixture similarity when combined with MoCo \cite{he2020momentum}, while we learn the similarity between the mixture and its parents (forming spurious-positive pairs). This poses a fundamental difference as the loss has to be redesigned accordingly. Notice \cite{shen2020rethinking} also designs a too complex approach to strive for semantical harmony by decaying the `context' image while not necessary in our case.

\begin{figure}[ht]
	\centering
	\includegraphics[width=0.6\columnwidth]{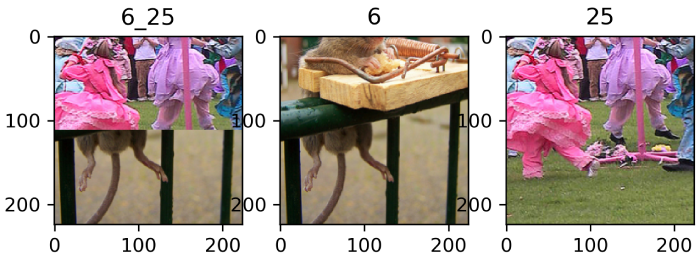}
	 \vskip 0.1in
	\caption{CutMix mixture for image 6 and 25 of Fig.~\ref{fig:mixup-32}. SIM is more sensitive to large color change.}
	\label{fig:mixup-6-26}
\end{figure}

\begin{figure}[ht]
	\centering
	\includegraphics[width=0.75\columnwidth]{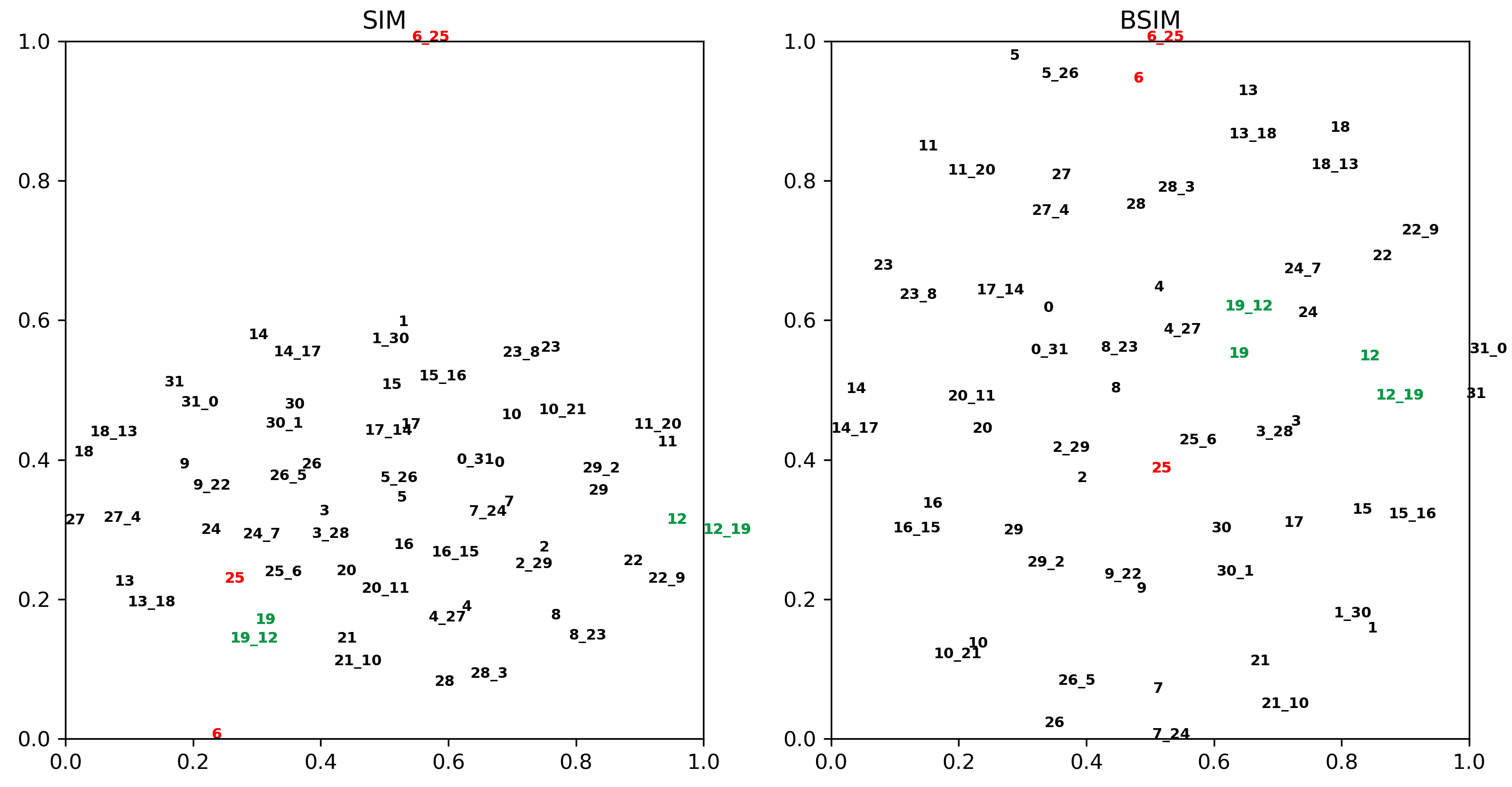}
	\vskip 0.1in
	\caption{The representation of 2D embeddings using TSNE. We use 32 images and construct the mixture using $i$-th and $31-i$-th images. The mixed embedding is marked by $i\_31-i$. Without BSIM, the embedding cannot make good use of the space (i.e. more crowded). \textbf{Left}: SIM, \textbf{Right}: BSIM. Notice SIM pushes a simple mixture $\textcolor{red}{6\_{25}}$ ({\color{red}{red}}) distant from others, leaving a narrow space and a large amount of unused space. Besides, $\textcolor{darkgreen}{12\_19}$ and $\textcolor{darkgreen}{19\_12}$ ({\color{darkgreen}{green}}) should be close but are pushed too far. In contrast, in BSIM all instances are scattered quite evenly, while mixed instances are also better placed near its parents. }
	\label{fig:mixup-32}
	 \vskip -0.1in
\end{figure}

\paragraph{TSNE Visualization to showcase the working mechanism of BSIM}
We extend the discussions about the working mechanism of BSIM. As mentioned Sec~\ref{sec:discuss} (main text), BSIM's latent space is less crowded to facilitate discrimination. To better illustrate this benefit, we pick 32 images and construct their mixtures and map their latent representations via TSNE, see Figure~\ref{fig:mixup-32}. It turns out that BSIM works like a ruler that measures how far a mixed instance should be from its parents. For instance, $6\_25$ (Figure~\ref{fig:mixup-6-26}) is distant from both 6 and 25 in SIM, while it is closer to 6 in BSIM. It is also evident that the latent space is evenly spaced in BSIM than SIM. Recall that TSNE is a dimension reducing tool, whose visualized distance is a relative measure in the
latent space, not necessarily proportional to its crop size.
Hence $6\_25$ shall not be accurately centered in between 6
and 25 although only each half of 6 and 25 are used for the
mixture. For SIM, $25\_6$ is close to 25 but $6\_25$ is far from
both. For BSIM, $6\_25$ is close to 6, but so is $25\_6$ to 25.
This subtle difference exactly manifests the difference of
the two. That is, we stretch out the latent space in terms of
the relative distance of every two instances to their spurious
pairs, while SIM can not.

Nevertheless, it is easier for decision making when instances are well-organized other than cluttered. The mixture near decision boundary can serve as a pivot for quantitively separating instances, which also helps scattering the representations evenly in the latent space.

\section{List of Additional Figures}
\begin{figure}[ht]
	\centering
		\includegraphics[width=0.4\columnwidth]{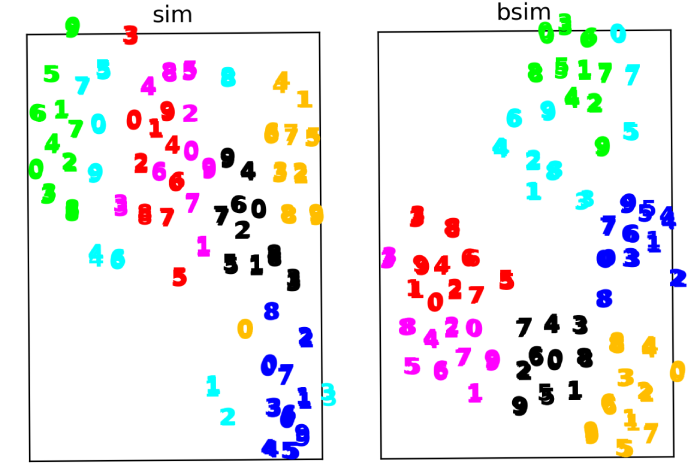}
		 \vskip 0.1in
	\caption{The representation of 2D embeddings using TSNE (sampled 3 times). We draw 10 instances (denoted by number) from each class and make 5 times of data augmentation per instance. The same class are denoted by the same color. BSIM has a better inter-class discrimination than SIM, where instances of the same class are mostly distant from those of other classes. }
	\label{fig:bsim-intra}
	\vskip -0.1in
\end{figure}

Figure~\ref{fig:bsim-intra} demonstrates that BSIM has better intra-class discrimination that SIM.

\begin{figure}[ht]
	\centering
	\includegraphics[width=0.6\columnwidth]{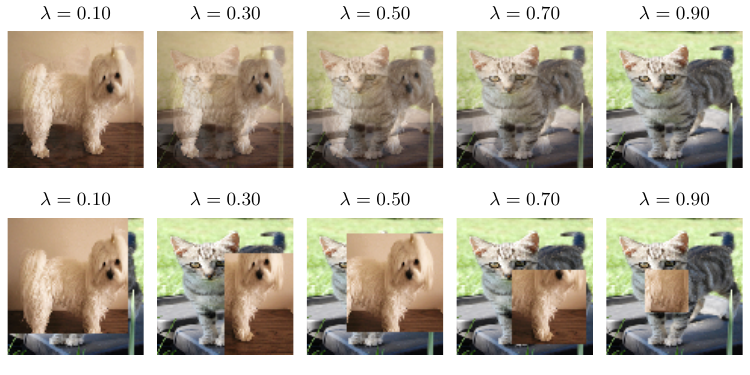}
	 \vskip 0.1in
	\caption{Comparison of Mixup and CutMix with $\lambda$ as the interpolation ratio for Mixup, and cutout ratio for CutMix}
	\label{fig:mixup-cutmix}
\end{figure}

Figure~\ref{fig:mixup-cutmix} shows the difference of mixed images between Mixup and CutMix, where the latter is perceptually more natural.

\begin{figure}[ht]
\centering
\includegraphics[width=0.5\columnwidth]{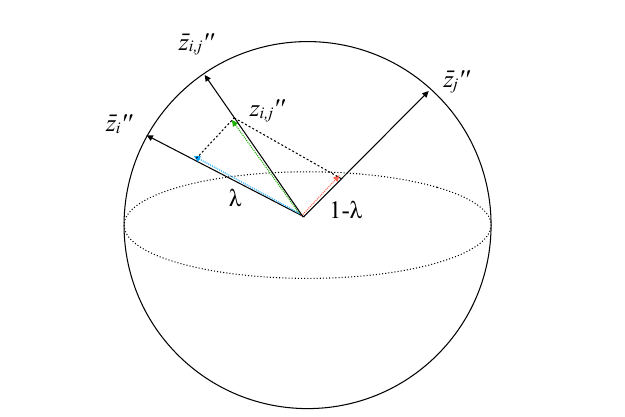}
\vskip 0.1in
\caption{Schematic view of the unit ball in high-dimensional space. The target embedding of the mixed image is  $\bar{z}''_{i,j}$, which is formed by  normalizing  $z_{i,j}''$ (dashed in green) to the sphere of the unit ball.  $z_{i,j}''$  is obtained by the $\lambda$ controlled linear  interpolation  between  $z_{i}''$ and  $z_{j}''$. }
\label{fig:unit-ball}
\end{figure}

Figure~\ref{fig:unit-ball} illustrates a schematic view of latent sphere where the mixed representation is normalized on the surface.

\begin{figure}[!t]
\small
\centering
\begin{tikzpicture}
    \node at (-0.9,1.5) (r1) {$\leftarrow\,$Representation$\,\rightarrow$};
    \node at (2.3,1.5) (r2) {$\leftarrow\,$Representation$\,\rightarrow$};
    \node[draw, circle, minimum size=26pt] at (-1,-1.5) (xi) {$\bm{x}_i$};
    \node[draw, circle, minimum size=26pt] at (2.5,-1.5) (xj) {$\,~\bm{x}_j$};
    \node[draw, circle, minimum size=26pt] at (-2.5,0) (xi2) {$\bm{x}_i''$};
    \node[draw, circle, minimum size=26pt] at (4,0) (xj2) {$\bm{x}_j''$};
    \node[draw, circle, minimum size=26pt] at (0.7,0) (xij) {$\bm{x}_{i,j}'$};
    \node at (-2.5,1.5) (hi2) {$\bm h_i''$};
    \node at (0.7,1.5) (hi) {$\bm h_{i,j}'$};
    \node at (-2.5,2.5) (zi2) {$\bm z_i''$};
    \node at (0.7,2.5) (zi) {$\bm z_{i,j}'$};
    \node at (4,1.5) (hj2) {$\bm h_j''$};
    \node at (4,2.5) (zj2) {$\bm z_j''$};
    \path[->] 
        (xi)  edge [>=latex] node[left,rotate=0] {$t''\sim\mathcal{T}''$} (xi2)
        (xi) edge [>=latex] node[left,rotate=0] {$\lambda, t'$} (xij)
        (xj) edge [>=latex] node[right,rotate=0] {$1-\lambda, t'$} (xij)
        (xj) edge [>=latex] node[right,rotate=0] {$t''\sim\mathcal{T}''$} (xj2)
        (xi2)  edge [>=latex] node[left,rotate=0] {$f(\cdot)$} (hi2)
        (hi2)  edge [>=latex] node[left,rotate=0] {$g(\cdot)$} (zi2)
        (xij)  edge [>=latex] node[right,rotate=0] {$f(\cdot)$} (hi)
        (hi)  edge [>=latex] node[right,rotate=0] {$g(\cdot)$} (zi)
        (xj2)  edge [>=latex] node[right,rotate=0] {$f(\cdot)$} (hj2)
        (hj2)  edge [>=latex] node[right,rotate=0] {$g(\cdot)$} (zj2);
    \path[<->]
        (zi2)  edge [>=latex] node[above,rotate=0] {$\mathrm{sim}$} (zi)
        (zj2)  edge [>=latex] node[above,rotate=0] {$\mathrm{sim}$} (zi);
    \end{tikzpicture}
    \vskip 0.1in
    \caption{Adapting BSIM into SimCLR for contrastive learning. Given two images $x_i$ and $x_j$ (we use $j=N-i$ for speed-up within a batch of N samples), we generate a spurious-positive sample by mixing separately augmented features ($t'\sim\mathcal{T}'$) into $x_{i,j}'$, which pairs with  $x_i''$ and $x_j''$. SimCLR is a special case of SimCLR-BSIM when $\lambda$ is 0 or 1.
     Note $f(\cdot)$ is an encoder network and $g(\cdot)$ refers to a projection head, both are trained with the contrastive loss in Equation \ref{eq:loss}. The representation $\bm h$ is later used for downstream tasks.}
    \label{fig:bsim_simclr_framework}
\end{figure}
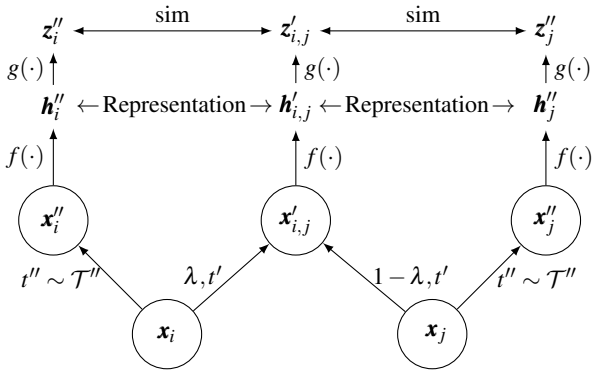

Figure~\ref{fig:bsim_simclr_framework} manifests the schematics of of SimCLR-BSIM.

\begin{figure}[ht]
\small
\centering
\begin{tikzpicture}[auto,node distance=0.9cm,semithick,scale=0.9, every node/.style={scale=0.9}]
\tikzstyle{var}=[draw, rectangle, minimum size=20pt]
\tikzstyle{arr}=[>=latex]
\node[var] at (-0.5,1) (xi) {$\bm x_i$};
\node[var] at (-0.5,-2) (xj) {$\bm x_j$};
\begin{scope}
\node[var] at (-0.5,0) (xi1) {$\bm x_i'$};
\node[var] at (-0.5,-1) (xj1) {$\bm x_j'$};
\node[var] at (1,-0.5) (xij1) {$\bm x_{i,j}'$};
\node[var] at (2.5,-0.5) (yt) {$\bm y_{\theta}'$};
\node[var] at (4,-0.5) (zt) {$\bm z_{\theta}'$};
\node[var] at (5.5,-0.5) (qt) {$\bm q_{\theta}(z_{\theta}')$};
\end{scope}
\begin{scope}
\node[var] at (1,1) (xi2) {$\bm x_i''$};
\node[var] at (2.5,1) (yk) {$\bm y_{i,\xi}''$};
\node[var] at (4,1) (zk) {$\bm z_{i,\xi}''$};
\end{scope}
\begin{scope}
\node[var] at (1,-2) (xj2) {$\bm x_j''$};
\node[var] at (2.5,-2) (yk2) {$\bm y_{j,\xi}''$};
\node[var] at (4,-2) (zk2) {$\bm z_{j,\xi}''$};
\end{scope}
\node[var, fill=green!20] at (5.5,1) (li) {$\mathcal{L}_{i,\theta,\xi}'$};
\node[var, fill=green!20] at (5.5,-2) (lj) {$\mathcal{L}_{j,\theta,\xi}'$};
\node[var, fill=green!20] at (7,-0.5) (loss) {$\mathcal{L}_{\theta, \xi}'$};
 \path[->] 
 	(xi)  edge [arr] node[left,rotate=0] {$t'$} (xi1)
        (xi)  edge [arr] node[above,rotate=0] {$t''$} (xi2)
        (xj)  edge [arr] node[left,rotate=0] {$t'$} (xj1)
        (xj)  edge [arr] node[above,rotate=0] {$t''$} (xj2)
        (xi1)  edge [arr] node[above,rotate=0] {$\lambda$} (xij1)
        (xj1)  edge [arr] node[below,rotate=0] {$1-\lambda$} (xij1)
        (xij1) edge [arr] node[above] {$f_{\theta}$} (yt)
        (xi2) edge [arr] node[above] {$f_{\xi}$} (yk)
        (xj2) edge [arr] node[above] {$f_{\xi}$} (yk2)
        (yt) edge [arr] node[above] {$g_{\theta}$} (zt)
        (yk) edge [arr] node[above] {$g_{\xi}$} (zk)
        (yk2) edge [arr] node[above] {$g_{\xi}$} (zk2)
        (zt) edge [arr] node[above] {$q_{\theta}$} (qt)
        (qt) edge [arr,dashed] (li)
        (qt) edge [arr,dashed] (lj)
        (zk) edge [arr,dashed](li)
        (zk2) edge [arr,dashed] (lj)
        (li) edge [arr] node[above,rotate=0] {$\lambda$}  (loss)
        (lj) edge [arr] node[right,rotate=0] {$1-\lambda$} (loss);
\begin{pgfonlayer}{background}
    \filldraw [line width=4mm,join=round,blue!10]
      (qt.north  -| qt.east)  rectangle (xij1.south  -| xij1.west);
     \filldraw [line width=4mm,join=round,red!10]
      (zk.north -| zk.east) rectangle (xi2.south -| xi2.west);
      \filldraw [line width=4mm,join=round,red!10]
      (zk2.north -| zk2.east) rectangle (xj2.south -| xj2.west);
  \end{pgfonlayer}
\end{tikzpicture}
 \vskip 0.1in
\caption{Applying BSIM to BYOL as in Equation~\ref{eq:cosine-lossv0}. The above blue region is the online network, the below red one is the same target network.}
\label{fig:byol-bsim-lossv2}
\end{figure}
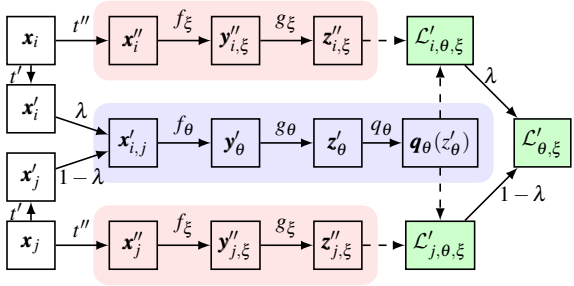

Figure~\ref{fig:byol-bsim-lossv2} gives the second implementation of BYOL-BSIM.

\begin{figure}[ht]
\centering
\includegraphics[width=0.6\columnwidth]{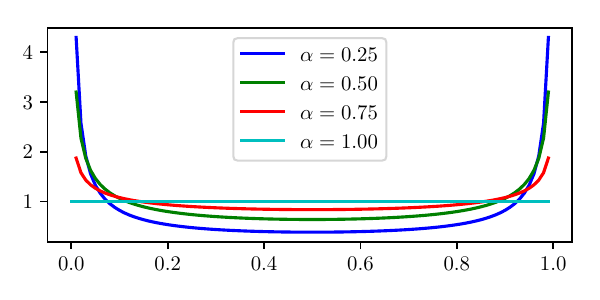}
 \vskip 0.1in
\caption{Probability density function of Beta distribution $\beta (\alpha, \alpha)$ under different settings. Notice when $\alpha=1$ we have a uniform distribution.}
\label{fig:best-dist}
\end{figure}

Figure~\ref{fig:best-dist} depicts the beta distribution given different $\alpha$, where we choose $\alpha$ to have a uniform distribution.

%
%

\begin{figure}[!t]
	\centering
	\includegraphics[width=0.4\columnwidth]{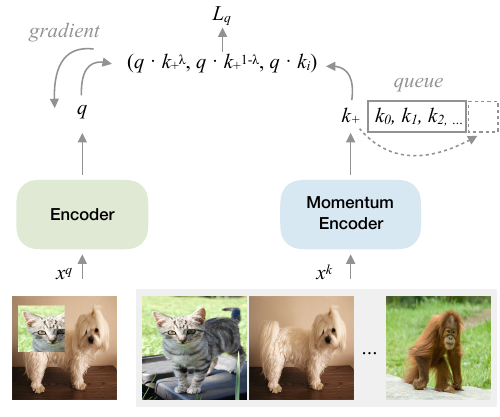}
	 \vskip 0.1in
	\caption{Applying BSIM to MoCo. Given a batch of images, we mix them in a pair-wise manner to produce $x_q$ (only one mixture is shown for simplicity).  We encode the mixed images as query $q$ and the whole current batch as keys $k$. We thus generate spurious-positive pairs from $q$ and $k$, and the current queue is used as negative samples.  
	}
	\label{fig:moco-bsim-framework}
\end{figure}

\clearpage

\end{document}